\documentclass[lettersize,journal]{IEEEtran}
\usepackage{amsmath,amsfonts}
\usepackage{algorithmic}
\usepackage{algorithm}
\usepackage{array}
\usepackage[caption=false,font=normalsize,labelfont=sf,textfont=sf]{subfig}
\usepackage{textcomp}
\usepackage{stfloats}
\usepackage{url}
\usepackage{hyperref}
\usepackage{verbatim}
\usepackage{graphicx}
\usepackage{cite}
\usepackage{threeparttable}
\usepackage{multirow}
\usepackage{dsfont}
\usepackage{makecell}
\usepackage{booktabs}
\usepackage{soul, color, xcolor}
\soulregister\cite7
\soulregister\ref7 
\newcommand{\zyq}[1]{\textcolor{red}{#1}}
\newcommand{\zzz}[1]{\textcolor{blue}{#1}}
\newcommand{\shan}[1]{\textcolor{black}{#1}}
\usepackage{hyperref} 

\begin{document}

\title{FOTS: A Fast Optical Tactile Simulator for Sim2Real Learning of Tactile-motor Robot Manipulation Skills}

\author{Yongqiang Zhao, Kun Qian\IEEEauthorrefmark{1}, Boyi Duan, Shan Luo 
\thanks{This paper has been accepted IEEE Robotics and Automation Letters.}
\thanks{This work is sponsored by Zhejiang Lab (No.2022NB0AB02), the Natural Science Foundation of Jiangsu Province (No.BK20201264), the National Natural Science Foundation of China (No.61573101), and \shan{the EPSRC project “ViTac: Visual-Tactile Synergy for Handling Flexible Materials” (EP/T033517/2).} \emph{(Corresponding author: Kun Qian)}}
\thanks{Yongqiang Zhao, Kun Qian, and Boyi Duan are with the School of Automation, Southeast University and the Key Laboratory of Measurement and Control of CSE, Ministry of Education, No.2, Sipailou, Nanjing 210096, China. E-mail: kqian@seu.edu.cn}
\thanks{Shan Luo is with the Department of Engineering, King's College London, London, WC2R 2LS, United Kingdom. E-mail:shan.luo@kcl.ac.uk}

}

\markboth{IEEE ROBOTICS AND AUTOMATION LETTERS}%
{}

\maketitle
\pagestyle{empty}  
\thispagestyle{empty} 
\begin{abstract}
Simulation is a widely used tool in robotics to reduce hardware consumption and gather large-scale data. Despite previous efforts to simulate optical tactile sensors, there remain challenges in efficiently synthesizing images and replicating marker motion under different contact loads. In this work, we propose a fast optical tactile simulator, named FOTS, for simulating optical tactile sensors. We utilize multi-layer perceptron mapping and planar shadow generation to simulate the optical response, while employing marker distribution approximation to simulate the motion of surface markers caused by the elastomer deformation. Experimental results demonstrate that FOTS outperforms other methods in terms of image generation quality and rendering speed, achieving 28.6 fps for optical simulation and 326.1 fps for marker motion simulation on a single CPU without GPU acceleration. In addition, we integrate the FOTS simulation model with physical engines like MuJoCo, and the peg-in-hole task demonstrates the effectiveness of our method in achieving zero-shot Sim2Real learning of tactile-motor robot manipulation skills. Our code is available at \href{https://github.com/Rancho-zhao/FOTS}{https://github.com/Rancho-zhao/FOTS}.
\end{abstract}

\begin{IEEEkeywords}
Tactile Sensing, Simulation, Robot Manipulation Skill, Tactile-motor Policy Learning, Sim-to-Real Transfer.
\end{IEEEkeywords}

\section{Introduction}
\IEEEPARstart{S}{imulation}  is increasingly important in the field of robotics. Compared to costly and time-consuming robot experiments in the real world, simulation has the potential to reduce wear and tear on robots and sensors, while providing a safer and more efficient alternative to generate data during policy exploration. However, simulating tactile sensors remains a challenging task that is essential for learning robot manipulation skills, even though several approaches and frameworks \cite{sferrazza2019ground, coumans2016pybullet, todorov2012mujoco} have been employed in robot simulation.

\begin{figure}[t]
	\centering
	\includegraphics[scale=0.35]{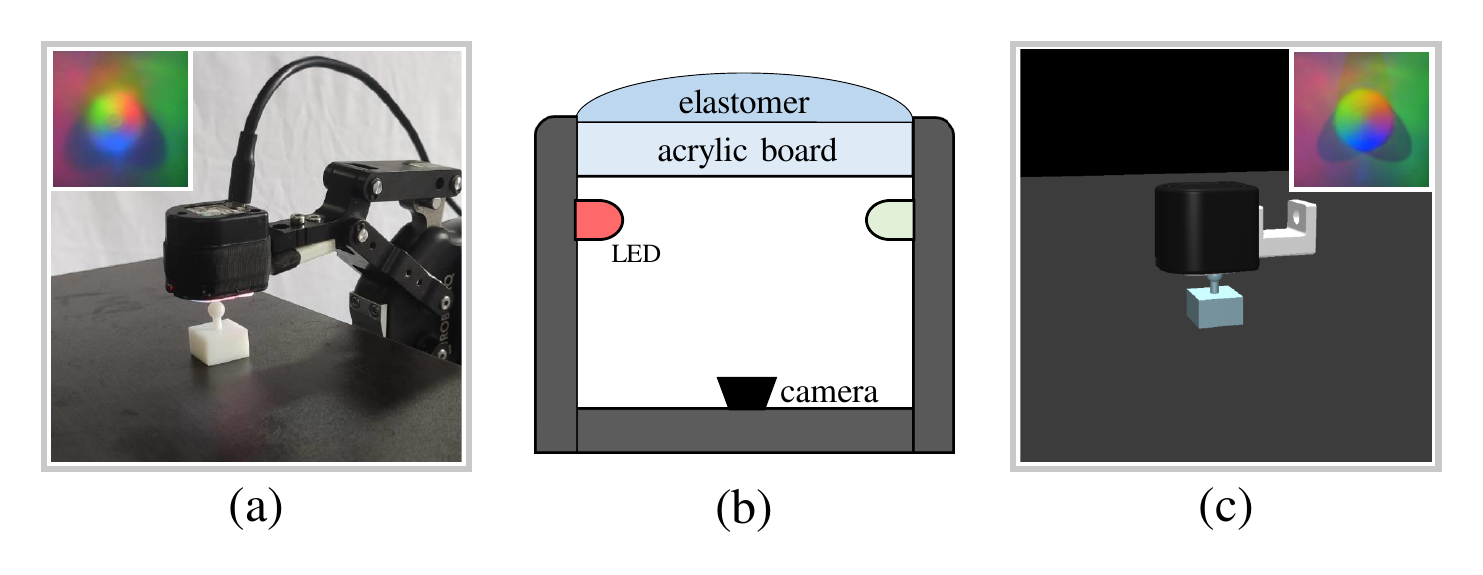}
	\caption{DIGIT sensor contacted with a spherical indenter. (a) A real-world sensor, (b) A section of the DIGIT sensor, (c) Our tactile simulator connected with MuJoCo.}\label{fig.real_sim}
\end{figure}

Robots, similar to humans, can benefit from utilizing tactile sensing when interacting with their environments to compensate for the limitations of visual sensing. In recent years, low-cost, high-resolution optical tactile sensors \cite{yuan2017gelsight, lambeta2020digit, gomes2020geltip} have been extensively applied in tactile-motor robot manipulation skills\cite{luo2018vitac, church2022tactile, pecyna2022visual}. As shown in Figure \ref{fig.real_sim}, DIGIT \cite{lambeta2020digit} is an example of such a sensor that uses an elastomer gel as the contact medium to interact with the environment. Deformable elastomer provides detailed and accurate tactile cues, but it also poses challenges in modeling the optical and mechanical responses of vision-based tactile sensors.

Despite the challenges of simulating deformable elastomers, numerous studies \cite{ding2020sim, agarwal2021simulation, gomes2021generation, wang2022tacto, si2022taxim, xu2023efficient, chen2023tacchi} have attempted to simulate different components of vision-based tactile sensors. Taichi-based models \cite{chen2023tacchi} focus on elastomer deformation during contact simulations so as to capture the mechanical characteristics of vision-based tactile sensors. However, these methods often come with significant computational costs. To enhance simulation speed, example-based models \cite{si2022taxim} and penalty-based models \cite{xu2023efficient} have been proposed. Example-based models show promise but still suffer from high computational costs associated with Finite Element Method (FEM) calibration. In contrast, penalty-based models lack accurate optical simulation. Previous works fall short in accurately representing optical or mechanical responses in complex contact scenarios under varying loads. Consequently, these limitations hinder the effective learning of tactile-motor robot manipulation skills using large-scale data \cite{sun2023learning}.

To this end, we propose FOTS, a Fast Optical Tactile Simulator, to synthesize realistic signals of light, shadow, and marker motion fields under different contact loads while accounting for intrinsic sensor noise. FOTS offers efficient computation and requires only tens of real-world tactile images collected during ball contact for calibration. Furthermore, we leverage FOTS to enhance tactile-motor policy learning in reinforcement learning tasks. Our zero-shot Sim2Real experiments showcase the effectiveness of FOTS in facilitating the development of tactile-motor robot manipulation skills by reducing the gaps between reality and simulation significantly.

In summary, the contributions of this work are three-fold:

\begin{itemize}
	\item We introduce a new multi-layer perceptron mapping method to simulate the optical response of optical tactile sensors by mapping contact gradients to pixel intensity in tactile images, and a planar shadow generation approach to replicate the shadows caused by point or directional lights, enhancing the realism of the optical tactile sensor simulation.
	\item We propose a novel approximation method that uses non-linear elastomer deformation to model the marker motion field, which outperforms other tactile simulators under various contact loads in terms of tactile flow generation quality and rendering speed.
	\item We apply the model to both DIGIT and GelSight sensors, as well as the tactile-motor peg-in-hole task. The easy-calibration and high-fidelity capabilities of FOTS facilitate integration across different sensors and enable efficient Sim2Real transfer of the learned policies.
\end{itemize}


\section{Related Work}
\label{SEC:2}
\subsection{Optical Simulation for Tactile Sensors}
Tactile sensing provides rich information about the states and properties of the contacted objects, such as shape, and texture \cite{luo2018vitac, kelestemur2022tactile}. Various types of tactile sensors based on different working principles have been developed \cite{luo2017robotic, nozaki2021development}, including resistive, capacitive, ultrasonic, magnetic, piezo-electric, and optical sensors. Among them, vision-based tactile sensors like GelSight \cite{yuan2017gelsight} and DIGIT \cite{lambeta2020digit} offer high-resolution tactile images. However, their soft elastomer is brittle and prone to wear during heavy contact. Therefore, generating vision-based tactile information in simulation is a feasible way to reduce real-world cost.

Several approaches have been proposed to simulate optical tactile sensors. Gomes et al. introduced a simulation method based on the Phong's model \cite{gomes2021generation}, which has been recently extended to accommodate curved sensors \cite{gomes2023beyond}. TACTO \cite{wang2022tacto} utilized Pyrender and bridged it with PyBullet to achieve realistic lighting and shadows for the DIGIT sensor. However, the above methods lack accurate calibration for simulating real-world sensors. The methods using Generative Adversarival Networks (GAN) \cite{jianu2022reducing,jing2023unsupervised} were able to synthesize tactile images, but incurred high training costs and required large datasets. Taxim \cite{si2022taxim} employed polynomial table mapping and unit shadow accumulation to generate simulated tactile images but lacked the ability to simulate non-flattened shadows. In contrast, we use multi-layer perceptron mapping and planar shadow generation to simulate the intrinsic lighting and shadows of real sensors. This not only achieves low computational costs but also ensures acceptable accuracy in the simulation process.

\subsection{Marker Motion Simulation for Tactile Sensors}
Optical tactile sensors can provide valuable information about the type and magnitude of contact force by utilizing printed marker patterns on or within the elastomer medium, in addition to capturing high-resolution tactile images. The movement of markers array on optical tactile sensors is a result of the gel deformation, which can be tracked by the camera. However, simulating the dynamics of deformable bodies, including the motion of markers on optical tactile sensors, remains a challenging task with existing methods \cite{yin2021modeling}. Ding et al. \cite{ding2020sim} simulated the dynamics of TacTip’s soft membrane in Unity to extract marker motion, which was then applied to Sim2Real robot tasks. Other approaches \cite{church2022tactile,kim2023marker} employed GANs to translate between realistic tactile images and simulated depth maps for TacTip and GelSight sensors. Taxim \cite{si2022taxim} explicitly simulated marker motion fields using offline FEM and online superposition principle, which required a gelpad FEM model and no extensive training data. A penalty-based tactile model capable of simulating both normal and shear loads was developed in \cite{xu2023efficient}. In this paper, instead of analyzing physical contact models or training networks, we introduce a novel and efficient method that accurately approximates the marker motion by employing marker distribution functions for three different types of loads.

\section{Methods}
\label{SEC:3}
\subsection{Overview}

\begin{figure}[t]
	\centering
	\includegraphics[scale=0.41]{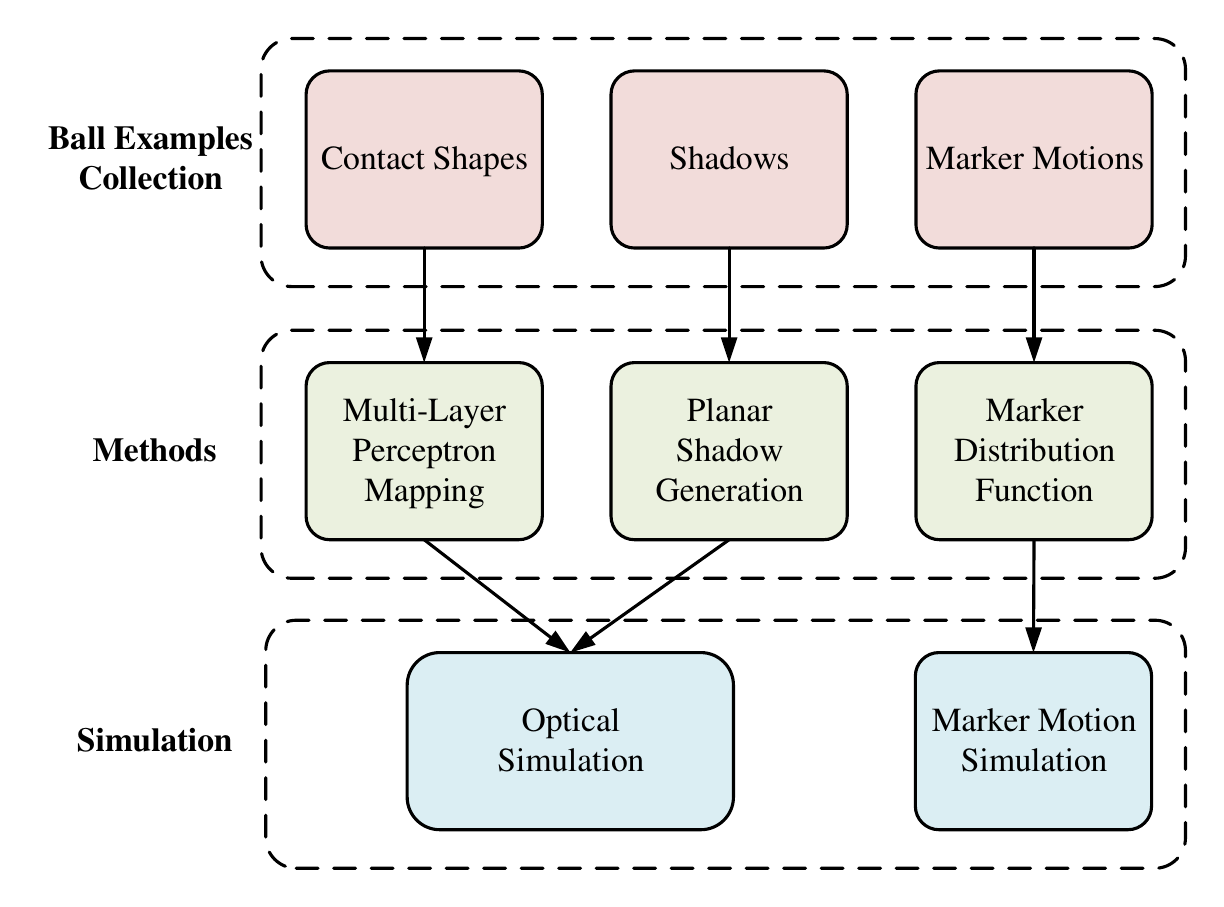}
	\caption{The pipeline of our proposed fast optical tactile simulator.}\label{fig.pipeline}
\end{figure}

The proposed method involves the construction of both optical and marker motion simulations as shown in Fig. \ref{fig.pipeline}. We map contact gradients to image intensities using multi-layer perceptron. The shadows for each light source are then simulated using planar shadow generation method, generating smoother edges than unit shadow generation approach in \cite{si2022taxim}. The marker distribution function is approximated for three different types of loads: normal, shear, and twist. This process enables the imitation of contact between objects and sensors in both static and dynamic scenarios in a realistic manner. In addition, each component of the simulator is calibrated with ball examples collected from real sensors.

\begin{figure}[t]
	\centering
	\includegraphics[scale=0.6]{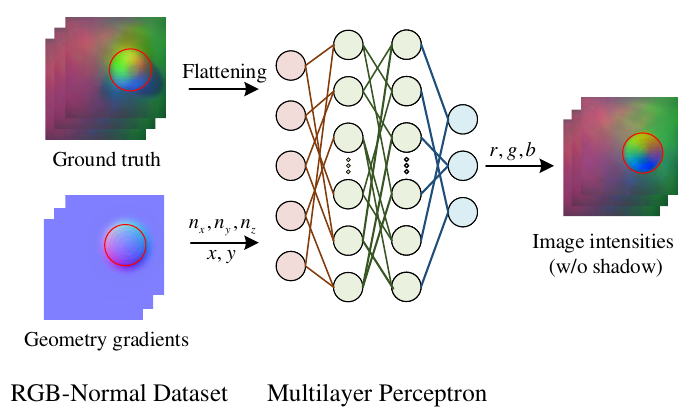}
	\caption{Mapping geometry gradients to image intensities using MLP.}
    \label{fig.rgb_normal}
\end{figure}

\subsection{Optical Simulation}
Following the assumption from \cite{yuan2017gelsight} that the reflection of light is evenly distributed, and the shading depends only on the local surface normal, we employ a multi-layer perceptron-based method to model the relationship between contact gradients and lighting intensities.

Given the height map $H$ generated from the gel surface of the tactile sensor, we can derive the light intensity $I$ based on the surface normal $\textbf{\textit{n}}$ at each point $P(x,y)$:

\begin{equation}
 \label{eq1}
I(x, y)=R\left(\frac{\partial H}{\partial x}, \frac{\partial H}{\partial y}\right)
\end{equation}
where $R$ is the reflectance function that characterizes both the lighting environment and the surface reflectance.

The reflectance function $R$ maps the geometry gradients to the image intensity, thereby enabling the lighting simulation of optical tactile sensors. However, the nonlinear relationship between $R$ and $\left(\frac{\partial H}{\partial x}, \frac{\partial H}{\partial y}\right)$ leads to the difficulty of modeling the light intensity. A feasible approach proposed in \cite{si2022taxim} is to apply polynomial table mapping, while we present a more robust method that utilizes a multi-layer perceptron (MLP) \cite{ramchoun2016multilayer} to model this relationship, as shown in Fig. \ref{fig.rgb_normal}. Our method uses real-world data to simulate the intrinsic noise of the real sensors, without requiring any prior knowledge of the sensors’ hardware layout. It also allows for greater generalization and robustness in the simulation, with an additional advantage of improved learning performance through the use of batch normalization. 

As the depth data obtained in the simulation provides the basis for the height map, we smooth the contact edge by convolving the height map with pyramid Gaussian kernels. By extracting the normal vector for each point on the height map, it is easy to map the vector to an intensity value using a trained MLP model, which is then utilized for tactile image synthesis.

\textbf{Calibration:} As Fig. \ref{fig.rgb_normal} shows, the calibration is performed by pressing a radius-known sphere indenter on the gel surface to capture tactile images. The data we use is each point's surface normal (calculated from the indenter’s geometry) and position within the circular area, as well as the point's RGB values as the corresponding annotation. With this data, we construct an RGB-Normal Dataset to train the MLP model. Note that we decouple lighting and shadows in this process, as shown in Fig. \ref{fig.rgb_normal}. The decoupling process is easy to achieve because of the clear boundary between lighting and shadows.

\subsection{Shadow Simulation}
In optical tactile sensors, the placement of internal light sources may cause shadows and changes in pixel intensity on tactile images when an object contacts the gel. Prior shadow simulation methods for optical tactile sensors\cite{wang2022tacto, si2022taxim} either rely on physical engine rendering, or are inefficient for non-flattened sensors (like DIGIT) due to curved surface calculation. Therefore, we use the planar shadow generation method \cite{woo1990survey} to eliminate platform dependencies and speed up shadow simulation, which can generate the shadow from each light source separately.

\begin{figure}[t]
	\centering
	\includegraphics[scale=0.15]{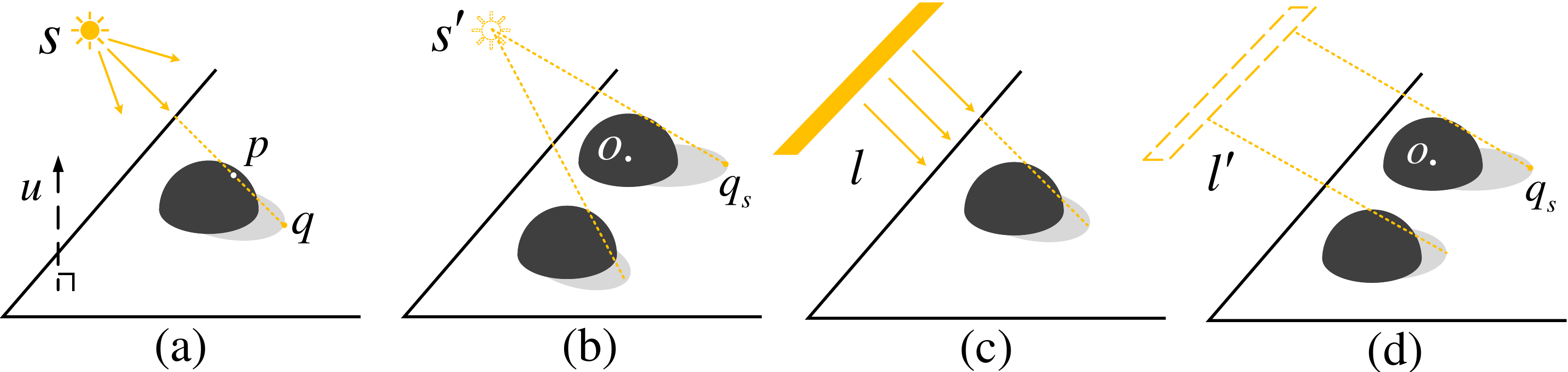}
	\caption{Sensor calibration by pressing a sphere indenter on the gel surface. (a) and (c) show planar shadow generation scenes from a point light, and a directional light, respectively; (b) and (d) show how to determine the position of a point light, and the direction of a directional light through multiple contact positions at different times, respectively.}
 \label{fig.shadow_sim}
\end{figure}

There are two main kinds of light sources used in optical tactile sensors: point (like DIGIT) and directional (like GelSight) lights. As shown in Fig. \ref{fig.shadow_sim}(a), for point light, we can calculate the shadow point $\textbf{\textit{q}}=(q_x,q_y,q_z)$ in the plane where the shadow is located, corresponding to an arbitrary point $\textbf{\textit{p}}=(p_x,p_y,p_z)$ of the contact model by the following equations:

\begin{equation}
 \label{eq2}
    \setlength{\arraycolsep}{1.0pt}
    \left[\begin{array}{c}
    q_{x}^{\prime} \\
    q_{y}^{\prime} \\
    q_{z}^{\prime} \\
    q_{w}^{\prime}
    \end{array}\right]=
    \left[\begin{array}{cccc}
    X & -u_{y} s_{x} & -u_{z} s_{x} & -d s_{x} \\
    -u_{x} s_{y} & Y & -u_{x} s_{y} & -d s_{y} \\
    -u_{x} s_{z} & -u_{y} s_{z} & Z & -d s_{z} \\
    -u_{x} & -u_{y} & -u_{z} & \textbf{\textit{u}} \cdot \textbf{\textit{s}}
    \end{array}\right]\left[\begin{array}{c}
    p_{x} \\
    p_{y} \\
    p_{z} \\
    1
    \end{array}\right]
\end{equation}
\begin{equation}
    X = \textbf{\textit{u}} \cdot \textbf{\textit{s}}+d-u_{x} s_{x}
\end{equation}
\begin{equation}
    Y = \textbf{\textit{u}} \cdot \textbf{\textit{s}}+d-u_{y} s_{y}
\end{equation}
\begin{equation}
    Z = \textbf{\textit{u}} \cdot \textbf{\textit{s}}+d-u_{z} s_{z}
\end{equation}
where $\left(\frac{q_{x}^{\prime}}{q_{w}^{\prime}}, \frac{q_{y}^{\prime}}{q_{w}^{\prime}}, \frac{q_{z}^{\prime}}{q_{w}^{\prime}}\right)=\left(q_{x}, q_{y}, q_{z}\right)$ represents the shadow point’s position, $\textbf{\textit{s}}=(s_x,s_y,s_z)$ represents the position of point light source, $\textbf{\textit{u}}=(u_x,u_y,u_z)$ is the unit normal vector of the plane where the shadow is located, and d is the distance parameter of the plane equation.

As for directional light, the equations are denoted as follows:
\begin{equation}
 \label{eq3}
    \setlength{\arraycolsep}{1.5pt}
    \left[\begin{array}{c}
    q_{x} \\
    q_{y} \\
    q_{z}
    \end{array}\right]=\frac{1}{\textbf{\textit{u}}\cdot \textbf{\textit{l}}}
    \left[\begin{array}{cccc}
    X^{\prime} & -u_{y} l_{x} & -u_{z} l_{x} & -d l_{x} \\
    -u_{x} l_{y} & Y^{\prime} & -u_{x} l_{y} & -d l_{y} \\
    -u_{x} l_{z} & -u_{y} l_{z} & Z^{\prime} & -d l_{z}
    \end{array}\right]\left[\begin{array}{c}
    p_{x} \\
    p_{y} \\
    p_{z} \\
    1
    \end{array}\right]
\end{equation}
\begin{equation}
    X^{\prime} = \textbf{\textit{u}} \cdot \textbf{\textit{l}}-u_{x} l_{x}
\end{equation}
\begin{equation}
    Y^{\prime} = \textbf{\textit{u}} \cdot \textbf{\textit{l}}-u_{y} l_{y}
\end{equation}
\begin{equation}
    Z^{\prime} = \textbf{\textit{u}} \cdot \textbf{\textit{l}}-u_{z} l_{z}
\end{equation}
where $\textbf{\textit{l}}=(l_x,l_y,l_z)$ represents the direction of directional light source.

Note that the depth reading in the simulation of the DIGIT sensor is not a fixed value due to its curved gel surface, as shown in Fig. \ref{fig.real_sim}(b). This brings about the difficulty in planar shadow generation. To address this issue, we subtract the in-contact depth map from a background tactile image captured with no object contact. Then we project the foreground of the depth map to a plane to yield a height map, which facilitates the generation of planar shadows. In detail, we set $\textbf{\textit{u}} = (0,0,1)$ and $d=0$ to optimize the computation. In addition, according to the specific light configuration of optical tactile sensors, we can choose a corresponding point or directional shadow generation method.

\textbf{Calibration:} According to Equations (\ref{eq2}) and (\ref{eq3}), the calibration process is to determine the position or the direction of each light source. As shown in Fig. \ref{fig.shadow_sim}, for point light, the convergence of rays provides a feasible calibration method to approximate the light source position by determining the nearest point of multiple tangents passing through corresponding points on the ball object and its oval shadow. The direction $\textbf{\textit{l}}'$ of the tangent can be calculated easily via tactile images contacted with a ball:

\begin{equation}
 \label{equ3}
    \textbf{\textit{l}}'=\{o_s,o_y,\| \textbf{\textit{o}} - \textbf{\textit{q}}_s \|_2 \cdot \tan(\text{arcsin}\frac{r}{\| \textbf{\textit{o}}-\textbf{\textit{q}}_s \|_2})\}
\end{equation}
where $r$ is the radius of the ball, $\textbf{\textit{o}}=(o_x,o_y,0)$ is the center of the ball, and $\textbf{\textit{q}}_s=(q_x,q_y,0)$ is the shadow vertex.

Based on $n$ tangents ($n\ge2$) and the least square method (LSM) we can obtain the nearest point of these tangents, which is the optimal light position. As for directional light, we can also obtain its direction $\textbf{\textit{l}}'$ via the above equation. In particular, the whole calibration only needs around 5 tactile images, which demonstrates that our shadow simulation method inspired by shadow generation in the real world is more efficient and low-cost than learning methods, like MLP.

\subsection{Marker Motion Simulation}
Marker motion on the gel surface is caused by the deformation of the elastomer from contacts. Instead of solving the inverse problem of estimating the force/torque from the marker motion \cite{si2022taxim, xu2023efficient}, we choose to model the marker displacement distribution function. This approach allows for direct approximation of the relationship between the marker motion and the contact geometry, whilst avoiding errors or inaccuracies caused by noise or calibration issues. Compared to the marker motion field simulation using force/torque information that requires precise measurements of gel deformation and marker displacement, our distribution modeling method is more robust to various factors, such as lighting, elastomer materials, and sensor alignment. In addition, due to the small size of the markers array, our method can achieve both high rendering speed and hight quality compared to neural networks such as MLP.

\begin{figure}[t]
	\centering
	\includegraphics[scale=0.4]{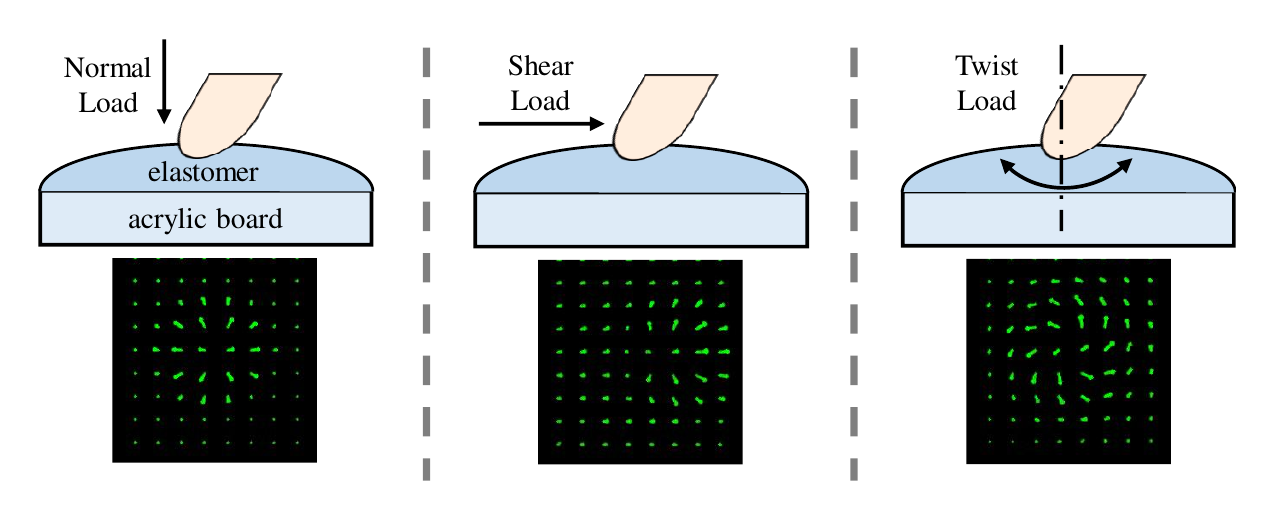}
	\caption{The markers motion of optical tactile sensors under different types of loads.}\label{fig.various_loads}
\end{figure}

In this work, we consider three types of marker motions corresponding to normal, shear, and twist loads, as Fig. \ref{fig.various_loads} shows. Our goal is to model the marker displacement distribution functions that can be employed in marker motions under different loads, and then combine them to simulate the deformation of the gel surface with contact loads. Although in this paper we apply our FOTS to simulate DIGIT and GelSight sensors, the marker motion simulation method is not limited to optical tactile sensors. Instead, it can be extended and applied to any other tactile sensors that utilize similar elastomer materials.

We model the marker displacement $\Delta d_d$ of the dilate motion resulting from the normal load and relative position between markers $M$ and the markers that is in contact $C_i$.
\begin{equation}
  \label{equ4}
  \Delta d_d=\sum_{i=1}^{N}\Delta h_i\cdot(M-C_i)\cdot\exp(-\lambda_d\|M-C_i)\|_2^2)
\end{equation}
where $N$ is the number of $C_i$; $\Delta h_i$ represents the height of $C_i$ relative to the initial gel surface when contacting objects; the direction of the marker displacement is determined via the term $M-C_i$; and the term $\exp(-\lambda_d\|M-C_i\|_2^2)$ represents the inverse relationship between the displacement of markers in motion and the distance from the load point. $\lambda_d$ is the coefficient of the exponential function that can affect the displacement caused by the dilate motion.

Similarly, the displacement of the markers $\Delta d_s$ under shear load and  $\Delta d_t$ under twist load can be modeled as:
\begin{equation}
  \label{equ5}
  \Delta d_s=\min\{\Delta s,\Delta s_{\max}\}\cdot\exp(-\lambda_s\|M-G\|_2^2)
\end{equation}
\begin{equation}
  \label{equ6}
  \Delta d_t=\min\{\Delta\theta,\Delta\theta_{\max}\}\cdot(M-G)\cdot\exp(-\lambda_t\|M-G\|_2^2) 
\end{equation}
where $G$ is a projection point of the object coordinate system origin on the gel surface along the normal direction of it. The first terms of the right sides of the two equations represent the effect of relative motion between objects and the gel surface. $\Delta s$ is the translation distance of $G$ relative to the gel surface coordinate system's origin, and $\Delta\theta=\begin{bmatrix}\cos\theta-1&-\sin\theta\\  \sin\theta&\cos\theta-1\end{bmatrix}$ is derived from $G$'s rotation degree $\theta$ combining the geometric relationships and trigonometric functions. $\Delta s_{\max}$ and $\Delta\theta_{\max}$ represent the conversion of friction from static to dynamic. The term $M-G$ can determine the direction of the marker displacement, and the last exponential terms represent the effect on the marker displacement caused by shear and twist loads. $\lambda_s$  and $\lambda_t$  are the coefficients of the exponential function. We determine these exponential functions by identifying the real marker motion's distribution under different loads as well as combining marker motion's generation principles, as Fig. \ref{fig.various_loads} shows.

The marker motion can be regarded as a result of these three loads, so we can simulate the marker motion combining these motion types:
\begin{equation}
  \label{equ7}
  M_c = M_{ini} + \Delta d_d + \Delta d_s + \Delta d_t
\end{equation}
where $M_c$ is the current positions of markers, and $M_{ini}$ is the initial positions of markers without object contact.

\textbf{Calibration:} We need to calibrate the values of $\lambda_d$, $\lambda_s$ and $\lambda_t$ for specific sensors to reduce the gap between reality and simulation.  Similar to the data collection in \cite{lei2022biomimetic}, we apply a sphere indenter to the sensor's gel surface to uniformly sample tens of points (15 points). For each point, the indenter is first pressed down a random depth to obtain the displacement data of the markers’ dilate motion. Then, we randomly shifted a distance around to obtain the displacement data of the marker’s shear motion (subtracting the dilate displacement caused by pressing). Finally, the indenter returns to the initial pressing position and randomly rotates by an angle to obtain the displacement data of the markers’ twist motion (also subtracting the dilate displacement). Based on these displacement functions and the collected data of markers’ dilate, shear, and twist motions, we utilize the nonlinear least square method (from lmfit python library) to fit $\lambda_d$, $\lambda_s$ and $\lambda_t$ separately. $\lambda_{d}=1.25e^{-3},\lambda_s=2.10e^{-4}$ and  $\lambda_t=3.80e^{-4}$ in our experiments for the DIGIT sensor.

\begin{figure*}[ht]
	\centering
	\includegraphics[scale=0.38]{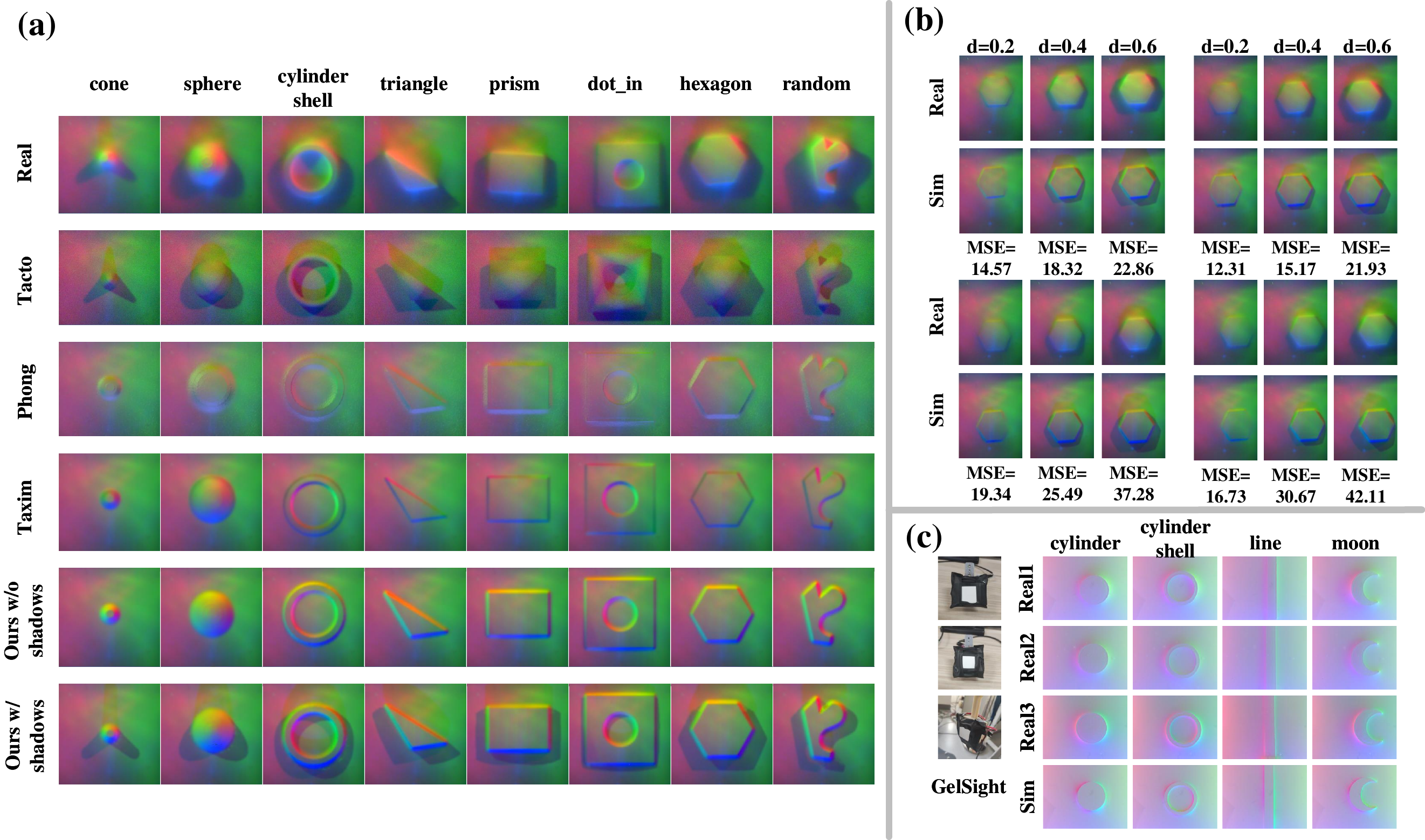}
	\caption{(a) The comparison of optical simulation among our method, TACTO\cite{wang2022tacto}, Phong\cite{gomes2021generation}, and Taxim\cite{si2022taxim} with the real data. (b) Optical simulation results with different indentation depths and locations. The locations differ over the gelpad surface while the depths differ by 0.2 mm, 0.4 mm, and 0.6 mm, repectively. The MSE error is shown below each pair. (c) Tactile images from GelSight sensors. The calibration data is from the first sensor "Real1".} \label{fig.optical_simulation}
\end{figure*}

\section{Experiments and Applications}
\label{SEC:4}
\subsection{Experiment Setup and Data Collection}
In this study, we connect FOTS with MuJoCo, but any physical engine like PyBullet and Gazebo with the capability to provide depth data and object positions can also be employed. Meanwhile, we set up a data collection platform (see Fig. \ref{fig.real_sim}(a)) to obtain the contact data of an indenter from a real DIGIT sensor mounted on a Robotiq 2F-85 gripper. The tool center point (TCP) of the robot can be adjusted manually to control the contact location and depth. In our previous work \cite{zhao2023skill}, we collected 2079 ($21\times 99$) real-world tactile images with 21 objects with different shapes from \cite{gomes2021generation} to form the Optical Dataset.

To calibrate the optical simulation model, we capture 50 tactile images from various locations on the gel surface. This was achieved by utilizing a 4mm-diameter spherical indenter. The images are then used to create an RGB-Normal Dataset consisting of 80,000 pixel points from the in-contact regions. For the shadow simulation model, we select 5 tactile images from the optical dataset. The calibration process is simple and takes less than an hour, without requiring precise control of contact locations. 

\subsection{Optical Simulation}

Based on the Optical Dataset and configuration settings, we synthesize the tactile images and compare our method with three other approaches: TACTO \cite{wang2022tacto}, Phong’s model \cite{gomes2021generation}, and Taxim \cite{si2022taxim}. We evaluate the performance of these methods by conducting a pixel-by-pixel comparison between the simulated and real-world images utilizing four metrics: mean absolute error (L1), mean squared error (MSE), structural index similarity (SSIM), and peak signal-to-noise ratio (PSNR). The simulated images we generate are $240\times 320$ pixels in size. Due to the accuracy of the operation with the real sensor, the ground truth tactile images do not match perfectly with the simulated images, promoting us to manually align the images using GIMP.

The qualitative results shown in Fig. \ref{fig.optical_simulation}(a) and the quantitative results summarized in Table \ref{Tab:2} both demonstrate that our method outperforms uncalibrated methods \cite{wang2022tacto,gomes2021generation} as well as data-driven method \cite{si2022taxim} similar to ours. For example, in the case of the “prism” images, our method excels in generating accurate shadows for the curved gel surface of DIGIT, whereas other methods fail.

On the other hand, we evaluate the performance of optical simulation across different scale-size training datasets. As Table. \ref{Tab:2} shows, the model's performance gradually improves with the number of tactile images increasing from 25 to 75. But, when the number exceeds 50, the performance improvement is not significant because of overfitting.

\textbf{Different indentation depths and locations:} The optical simulation model in our study demonstrates robust performance across various indentation depths and positions. As Fig. \ref{fig.optical_simulation}(b) shows, it is evident from MSE analysis that errors exhibit an upward trend as the indentation moves farther away from the central region and deepens. This comes from the less calibration data in areas far from the center and the approximation model of soft body deformation becomes less effective for large displacements.

\begin{figure*}[ht]
	\centering
	\includegraphics[scale=0.35]{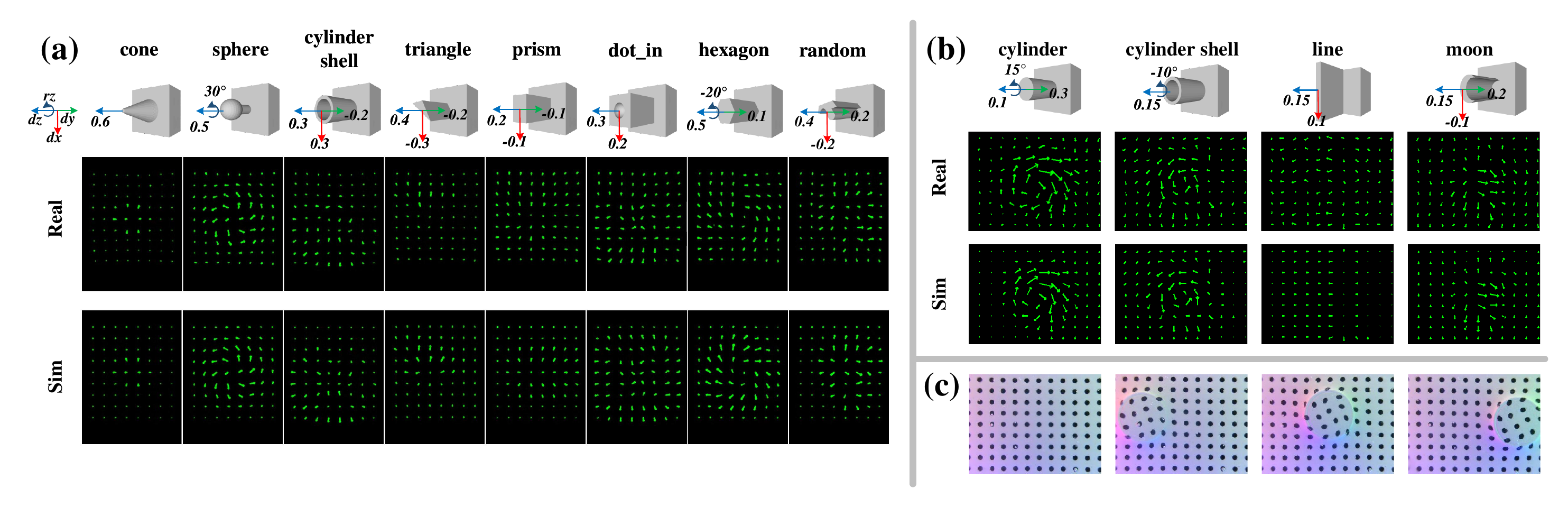}
	\caption{(a) Marker motion field simulation results of a DIGIT sensor. We visualize the marker motions (scaled up by 5 for better visualization) on the objects from \cite{gomes2021generation} under different normal, shear, and twist loads. (b) Marker simulation of a GelSight sensor whose marker arrangement is different from the DIGIT sensor. (c) shows marker distortion under different loads}\label{fig.marker_quality}
\end{figure*}

\begin{table}
	\centering
	\caption{Image similarity metrics indicating the gap from simulated tactile data to real-world tactile data}\label{Tab:2}
	\setlength{\tabcolsep}{3mm}{
	\begin{tabular}{ccccc}
	  \toprule
	  {Methods} & {L1 $\downarrow$} &{MSE $\downarrow$} &{SSIM $\uparrow$} &{PSNR $ \uparrow$} \\
      \midrule
       {TACTO\cite{wang2022tacto}} & {10.310} & {201.872} & {0.798} & {25.425} \\
       {Phong\cite{gomes2021generation}} & {8.263} & {125.329} & {0.830} & {28.278} \\
       {Taxim\cite{si2022taxim}} & {5.135} & {56.715} & {0.882} & {31.115} \\
       {Ours w/o shadows-50} & {5.032} & {55.223} & {0.876} & {31.241} \\
       {Ours w/ shadows-25} & {6.210} & {70.903} & {0.853} & {28.675} \\
       {Ours w/ shadows-50} & {4.864} & {52.451} & {\textbf{0.894}} & {32.587} \\
       {Ours w/ shadows-75} & {\textbf{4.861}} & {52.435} & {0.892} & {\textbf{32.620}} \\
       {Ours w/ shadows-100} & {4.867} & {\textbf{52.418}} & {\textbf{0.894}} & {32.613} \\
	  \bottomrule
	\end{tabular}}
	\vspace{-10pt}
\end{table}

\begin{table}
	\centering
	\caption{Image similarity metrics indicating the generalization across inter-class GelSight Sensors}\label{Tab:new}
	\setlength{\tabcolsep}{3mm}{
	\begin{tabular}{ccccc}
	  \toprule
	  {Sensors} & {L1 $\downarrow$} &{MSE $\downarrow$} &{SSIM $\uparrow$} &{PSNR $ \uparrow$} \\
      \midrule
       {Real1} & {4.912} & {53.267} & {0.886} & {32.297} \\
       {Real2} & {5.010} & {53.631} & {0.881} & {32.029} \\
       {Real3} & {5.153} & {54.317} & {0.872} & {31.103} \\
	  \bottomrule
	\end{tabular}}
	\vspace{-10pt}
\end{table}

\begin{table}
	\centering
	\caption{Speed test for optical simulation on CPU}\label{Tab:3}
	\setlength{\tabcolsep}{3mm}{
	\begin{tabular}{cccccc}
	  \toprule
	  \multirow{2}*{Methods} & {Ours w/o} &{Ours w/} &{Taxim} &{Phong} & {TACTO}\\
        &{shadows}&{shadows}&{\cite{si2022taxim}}&{\cite{gomes2021generation}}& {\cite{wang2022tacto}}\\
      \midrule
       {Speed(fps)} & {37.0} & {28.6} & {16.7} & {11.1} & {5.2} \\
	  \bottomrule
	\end{tabular}}
	\vspace{-10pt}
\end{table}

\textbf{Simulation on various sensors:} As Fig. \ref{fig.real_sim} shows, our initial application of the model involves the DIGIT sensor, yielding strong performance. Subsequently, we employ distinct GelSight sensors to further showcase the effectiveness of our method, as shown in Fig. \ref{fig.optical_simulation}(c) and Table \ref{Tab:new}. We collect calibration data from the first GelSight sensor "Real1". The second sensor "Real2" is manufactured using a nearly identical process to the first one, while the third sensor "Real3" differs in its shell design and light intensity. These images and metrics indicate our method's generalization capabilities across inter-class optical tactile sensors.

\textbf{Speed test:} All the aforementioned simulation models are tested on an Intel Core i7-9700k 8-Core Processor CPU. Using height maps of size $240\times 320$ as input, we generate the simulated tactile images as outputs and measure the average running time of each method, which is reported in Table \ref{Tab:3}. Our method proves to be the most efficient on CPU, achieving real-time data transfer speed from real sensors.

\subsection{Marker Motion Simulation}
We calibrate marker motion simulation by collecting 150 ($50\times 3$) tactile images from different locations on the gel surface using a 4mm-diameter spherical indenter for three types of loads. We evaluate the simulation results against the real-world marker displacement images captured from the above 21 objects that are in contact with the sensor’s gel surface under different combinations of normal, shear, and twist loads. Some examples of results are shown in Fig. \ref{fig.marker_quality}. Furthermore, we compare our method with the penalty-based model \cite{xu2023efficient} and FEM-based model \cite{si2022taxim} using three real-world objects (sphere, cylinder, and cuboid) under various loads ranging from 0.1mm to 0.6mm and -50° to 50°. We distinguish tactile flows (as Fig. \ref{fig.marker_quality} shows) from optical images to reduce the effect of marker distortion (as Fig. \ref{fig.marker_quality}(c) shows). Table \ref{Tab:4} demonstrates our method outperforms other methods in terms of the mean marker displacement L1 error between real-world and simulated results under normal and twist loads, which are $1.07e^{-2}$ mm and $1.53e^{-2}$ mm, respectively. Meanwhile, our method’s mean L1 error is close to other methods under shear loads, which is $1.29e^{-2}$ mm. 

As Fig. \ref{fig.marker_quality}(b) shows, our marker simulation method still performs well on another type of optical tactile sensor (i.e. the GelSight) with a different marker arrangement.

\begin{table}
	\centering
	\caption{The comparison of the mean L1 error (1e-2 mm) of markers displacement under loads}\label{Tab:4}
	\setlength{\tabcolsep}{3mm}{
	\begin{tabular}{cccc}
	  \toprule
	  {Methods} & {Normal $\downarrow$} & {Shear $\downarrow$} & {Twist $\downarrow$} \\
      \midrule
       {Taxim\cite{si2022taxim}} & {1.10} &{\textbf{1.28}}& {/} \\
       {\cite{xu2023efficient}'s Method} & {1.12} & {\textbf{1.28}} & {1.56} \\
       {Ours} & {\textbf{1.07}} & {1.29} & {\textbf{1.53}} \\
	  \bottomrule
	\end{tabular}}
	\vspace{-10pt}
\end{table}


\begin{figure*}[ht]
	\centering
	\includegraphics[scale=0.28]{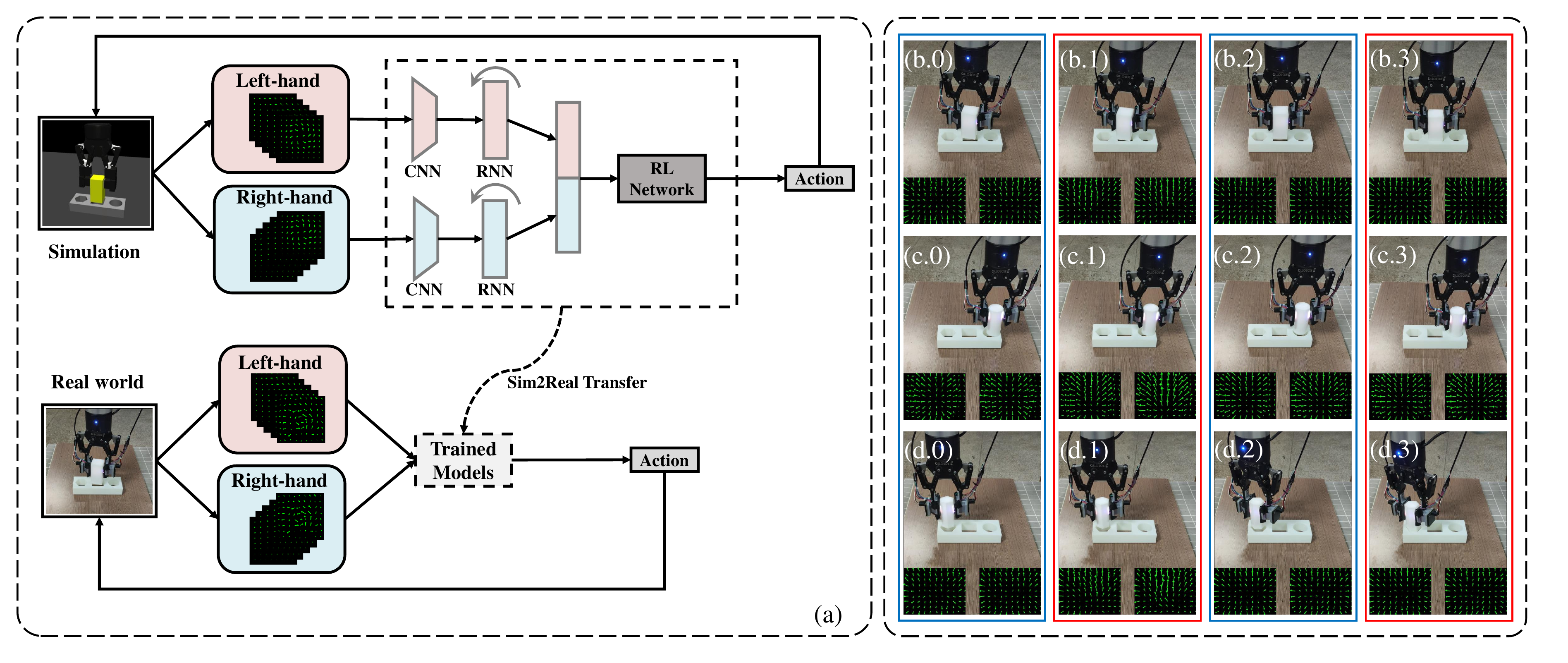}
	\caption{(a) The policy learning framework and Sim2Real learning of tactile-motor peg-in-hole tasks using GelSight sensors with markers. (b-d) The real-world experiment snapshots (including tactile flow images of sensors mounted to the gripper at the bottom of each snapshot) of cuboid, cylinder, and hexagon objects. \zzz{Blue boxes} show insertion pose adjustments, and \zyq{red boxes} show insertion attempts. The tactile flows are scaled up by 5 for better visualization. }\label{fig.tactile_policy}
\end{figure*}

\subsection{Application in Tactile-motor Manipulation Task}


In this work, we use the Soft Actor-Critic (SAC) \cite{haarnoja2018soft} algorithm to learn the peg-in-hole task to demonstrate our approach's effectiveness. As shown in Fig. \ref{fig.tactile_policy}, we apply a Sim-to-Real policy framework proposed in our previous work \cite{zhao2023skill}. As Fig. \ref{fig.tactile_policy} shows, we use tactile flows as observations. The action is composed of 2D translation ($x$-$y$ plane) and 1D rotation ($z$ axis) displacement of the Tool Center Point (TCP) of the end-effector in the world coordinate system: $A=[\Delta x, \Delta y, \Delta r_z]$.

\textbf{Rewards:} We perform this task based on a similar task in \cite{xu2023efficient}. As Fig. \ref{fig.tactile_policy} shows, this task involves controlling a gripper to insert an object into a predefined hole with the same shape. The object has randomly initialized both translation and rotation misalignment to the hole. The insertion process is modeled as an episodic policy that iterates between insertion attempts followed by insertion pose adjustments, and the maximum number of attempts we set is 15. The robot needs to recognize tactile flows’ differences between two fingers when the object contacts the hole in different poses to decide the insertion pose adjustment. The reward function is:

\begin{equation}
  \label{equ9}
  R = p_0 g_{diff} + p_1 a_{diff} + p_2 s_p
\end{equation}
where $g_{diff}$ represents the distance difference between the object and the hole, $a_{diff}$ represents the angle difference between the object and the hole, and $s_p$ is the reward of inserting the object into the hole successfully. The weights are $p_0=-10, p_1=-1, p_2=-5$ which are obtained through grid search in our experiments.

We employ the methods from \cite{si2022taxim, xu2023efficient}, and ours, to create simulated environments. These environment setups are identical, differing only in their respective tactile simulators. As Fig. \ref{fig.tactile_policy}(a) shows, we train the policies and zero-shot transfer them to the same real-world scene. The zero-shot Sim-to-Real performance of our tasks is shown in Table \ref{Tab:5}.

\begin{table}
	\centering
	\caption{Zero-shot sim-to-real performance of peg-in-hole task using different simulators}\label{Tab:5}
	\setlength{\tabcolsep}{0.6mm}{
	\begin{tabular}{ccccccc}
	  \toprule
       \multirow{2}*{Methods}& \multicolumn{2}{c}{Cuboid} &\multicolumn{2}{c}{Cylinder}& \multicolumn{2}{c}{Hexagon} \\
       & {Success} & {Attempts} & {Success} & {Attempts} & {Successs} & {Attempts} \\
      \midrule
       {Taxim\cite{si2022taxim}} & {44.0\%}&{6.13} &{60.0\%}&{5.28}& {40.0\%}&{6.57} \\
       {\cite{xu2023efficient}'s Method} & {62.0\%}&{5.01} &{74.0\%}&{4.25}& {60.0\%}&{5.42} \\
       {Ours}&{\textbf{72.0\%}}&{\textbf{4.25}} &{\textbf{80.0\%}}&{\textbf{3.84}}& {\textbf{66.0\%}}&{\textbf{4.89}} \\
	  \bottomrule
	\end{tabular}}
	\vspace{-10pt}
\end{table}

Fig. \ref{fig.tactile_policy}(b-d) show the snapshots of different objects (cuboid, cylinder, and hexagon), and the task is successful when the object is inserted into the hole. As Table \ref{Tab:5} shows, the performance of Taxim on this task is hindered by the absence of marker motion simulation under twist load. In contrast, compared to the method proposed by \cite{xu2023efficient} considering different types of loads, our marker motion simulation method employed for tactile flows demonstrates higher success rates (72.0\%, 80.0\%, and 66.0\%) and less average number of attempts (4.25, 3.84 and 4.89) for all three objects after conducting 50 experiments.

Due to its accurate simulation of optical and mechanical responses of tactile sensors, our method excels in Sim2Real transfer performance in tactile-motor manipulation tasks. However, there is still a gap between simulation and reality, which we attribute to two factors: 1) the difficulties of fully simulating the physical characteristics of contacted objects; 2) our method’s inability to model the marker motion completely, such as the motion under the tilt torque load that is common for real contact cases.

\section{Conclusion And Future Work}
\label{SEC:5}
In this paper, we propose FOTS, an optical tactile simulator, to synthesize optical signals under various loads at a speed faster than any state-of-the-art simulator. We simulate optical responses using multi-layer perceptron mapping and planar shadow generation while applying marker distribution approximation to simulate the motion of the markers. The image quality of the tactile images generated by FOTS outperforms existing methods, and it achieves high accuracy on marker motion field simulation. In addition, the tactile-motor peg-in-hole task in Sim2Real settings is successfully learned and zero-shot transferred, demonstrating the generalization and transferability of FOTS. In the future, we plan to integrate the force/torque module into our simulator to further reduce the Sim2Real gap.



\end{document}